\documentclass[10pt,journal,cspaper,compsoc]{IEEEtran}

\ifCLASSOPTIONcompsoc
\else
\fi

\ifCLASSINFOpdf
\else
\fi

\usepackage{cite}
\usepackage{makeidx} \makeindex
\usepackage{graphicx}
\usepackage{amsmath,amssymb} 
\usepackage{bm}
\usepackage{booktabs} 
\usepackage{multirow}
\usepackage{epstopdf}
\usepackage{url}
\usepackage{ulem}
\usepackage{color}

\newcommand{\eg}{\emph{e.g.}}
\newcommand{\ie}{\emph{i.e.}}
\graphicspath{{./Imgs/}}

\begin{document}

\title{Simple to Complex: A Weakly-supervised Framework for Semantic Segmentation}
\title{{STC: A Simple to Complex Framework for Weakly-supervised Semantic Segmentation}}

\author{Yunchao Wei,
		Xiaodan Liang,
		Yunpeng Chen,
		Xiaohui Shen,
		Ming-Ming Cheng,
		Jiashi Feng,
		Yao Zhao, ~\IEEEmembership{Senior Member,~IEEE} and
		Shuicheng Yan ~\IEEEmembership{Senior Member,~IEEE}

\thanks{Yunchao Wei and Yao Zhao are with the Institute of Information Science, Beijing Jiaotong University, China, e-mail: wychao1987@gmail.com; yzhao@bjtu.edu.cn. Xiaodan Liang is with Sun Yatsen University, China, e-mail: xdliang328@gmail.com. Xiaohui Shen is with Adobe Research, U.S., e-mail: xshen@adobe.com. Ming-Ming Cheng is with CCCE, Nankai University, Tianjin, China, e-mail: cmm@nankai.edu.cn. Yunpeng Chen, Jiashi Feng and Shuicheng Yan are with Department of Electrical and Computer Engineering, National University of Singapore, e-mail: qw.2080@gmail.com; elefjia@nus.edu.sg; eleyans@nus.edu.sg.
}
}

\markboth{IEEE TRANSACTIONS ON PATTERN ANALYSIS AND MACHINE INTELLIGENCE,~Vol.XX, No.XX,~2015}%
{Shell \MakeLowercase{\textit{et al.}}: Bare Demo of IEEEtran.cls for Computer Society Journals}
%


\IEEEcompsoctitleabstractindextext{%
\begin{abstract}
Recently, significant improvement has been made on semantic object segmentation 
due to the development of deep convolutional neural networks (DCNNs).
Training such a DCNN usually relies on a large number of images with pixel-level
segmentation masks,
and annotating these images is very costly in terms of both finance and human effort.
In this paper, we propose a simple to complex (STC) framework in which 
only image-level annotations are utilized to learn DCNNs for semantic segmentation.
Specifically, we first train an initial segmentation network called Initial-DCNN 
with the saliency maps of simple images 
(\ie, those with a single category of major object(s) and clean background).
These saliency maps can be automatically obtained by existing bottom-up salient object detection techniques, 
where no supervision information is needed.
Then, a better network called Enhanced-DCNN is learned with supervision from 
the predicted segmentation masks of
simple images based on the Initial-DCNN as well as the image-level annotations.
Finally, more pixel-level segmentation masks of complex images
(two or more categories of objects with cluttered background),
which are inferred by using Enhanced-DCNN and image-level annotations,
are utilized as the supervision information
to learn the Powerful-DCNN for semantic segmentation.
Our method utilizes $40$K simple images from \texttt{Flickr.com} 
and 10K complex images from PASCAL VOC for step-wisely boosting the segmentation network.
Extensive experimental results on PASCAL VOC 2012 segmentation benchmark well demonstrate the superiority of the proposed STC framework compared with other state-of-the-arts.

\end{abstract}

}

\maketitle

\IEEEdisplaynotcompsoctitleabstractindextext

%

\IEEEpeerreviewmaketitle

\section{Introduction}
\IEEEPARstart{I}{n} recent years, deep convolutional neural networks (DCNNs) 
have demonstrated an outstanding capability in various computer vision tasks, 
such as image classification
\cite{krizhevsky2012imagenet,simonyan2014very,szegedy2014going,wei2014cnn},
object detection~\cite{girshick2014rich,girshick15fastrcnn} and semantic segmentation
\cite{xia2013semantic,2015-dai,hariharan2014simultaneous,2015-zheng-conditional,2015-long,dai2015convolutional,chen2014semantic}.
Most DCNNs for these tasks rely on strong supervision for training, 
\ie, ground-truth bounding boxes and pixel-level segmentation masks.
However, compared with convenient image-level labels, 
collecting annotations of bounding boxes or pixel-level masks 
is much more expensive. 
In particular, for the semantic segmentation task, annotating a large
number of pixel-level masks usually requires a considerable amount 
of financial expenses as well as human effort.

To address this problem, some methods
\cite{2015-papandreou-weakly,pinheiro2015weakly,pathak2014fully,xu2015learning,pathak2015constrained}
have been proposed for semantic segmentation by only utilizing image-level labels 
as the supervised information.
However, to the best of our knowledge, 
the performance of these methods is far from satisfactory compared with
fully-supervised schemes (\eg, $40.6\%$~\cite{pinheiro2015weakly} \emph{vs.} $66.4\%$~\cite{chen2014semantic}).
Given the complexity of semantic segmentation problems, 
such as high intra-class variation (\eg, diverse appearance, viewpoints and scale) 
and different interaction between objects (\eg, partial visibility and occlusion),
complex loss functions (\eg, multiple instance learning based loss functions)~\cite{pinheiro2015weakly,pathak2015constrained,2015-papandreou-weakly} with image-level annotations may not be adequate
for weakly supervised semantic segmentation due to 
the ignorance of intrinsic pixel-level properties of segmentation masks.

It should be noted that, during the past few years, 
many salient object detection methods
\cite{2015-aliborji,cheng2015global,shi2016hierarchical,jiang2013salient}, 
which do not require high-level supervision information, 
have been proposed to detect the most visually noticeable salient object in the image.
While these methods may not work well for complex images with multiple objects 
and cluttered background, 
they often provide satisfactory saliency maps for
images with the object(s) of single category and clean background.
By automatically retrieving a huge amount of web images and 
detecting salient objects for relatively simple images,
we might be able to obtain a large amount of saliency maps 
for training semantic segmentation DCNNs at a low cost.

In this work, we propose a simple to complex framework for weakly-supervised segmentation based on the following intuitions.
For complex images with clutter background and two or more categories of objects, 
it is usually difficult to infer the relationship between semantic labels and pixels by only utilizing 
image-level labels as the supervision. However, for simple images with clean background and 
a single category of major object(s), foreground and background pixels are 
easily split based on the salient object detection 
techniques~\cite{itti1998model,cheng2015global,shi2016hierarchical,jiang2013salient}. With the indication of the image-level label, it is 
naturally inferred that pixels belonging to the foreground can be assigned with the same semantic label.
Therefore, an initial segmenter can be learned from simple images based on their foreground/background masks and 
image-level labels. Furthermore, based on the initial segmenter, more objects from complex images can be segmented 
so that a more powerful segmenter can be continually learnt for semantic segmentation.

Specifically, semantic labels are firstly employed as queries to retrieve images on the image hosting websites,
\eg, \texttt{Flickr.com}.
The retrieved images from the first several pages usually meet the definition of a simple image.
With these simple images, high quality saliency maps are generated by the state-of-the-art 
saliency detection technique~\cite{jiang2013salient}.
Based on the supervision of image-level labels, we can easily assign a semantic label to each foreground pixel
and learn a semantic segmentation DCNN supervised by the generated saliency maps by employing a multi-label
cross-entropy loss function, in which each pixel is classified to both the \emph{foreground} class and \emph{background}
according to the predicted probabilities embedded in the saliency map.
Then, a simple to complex learning process is utilized to gradually improve the capability of DCNN, in which the
predicted segmentation masks of simple images by initially learned DCNN are in turn used as the supervision to
learn an enhanced DCNN.
Finally, with the enhanced DCNN, more difficult and diverse masks from complex images are further utilized for learning a more powerful DCNN. Particularly, the contributions of this work are summarized as follows:
\begin{itemize}	
	\item[$\bullet$] We propose a simple to complex (STC) framework that can effectively train the
	segmentation DCNN in a weakly-supervised manner (\ie, only image-level labels are provided).
	The proposed framework is general, and any state-of-the-art fully-supervised network structure can
	be incorporated to learn the segmentation network.
	\item[$\bullet$] A multi-label cross-entropy loss function is introduced to train a segmentation network
	based on saliency maps, where each pixel can adaptively contribute to the \emph{foreground} class and \emph{background}
	with different probabilities.

	\item[$\bullet$] We evaluate our method on the PASCAL VOC 2012 segmentation benchmark~\cite{2010-pascal}.
	The experimental results well demonstrate the effectiveness of the STC framework, achieving the state-of-the-art performance.

\end{itemize}

\begin{figure}[t]
	\centering
	\includegraphics[scale=0.42]{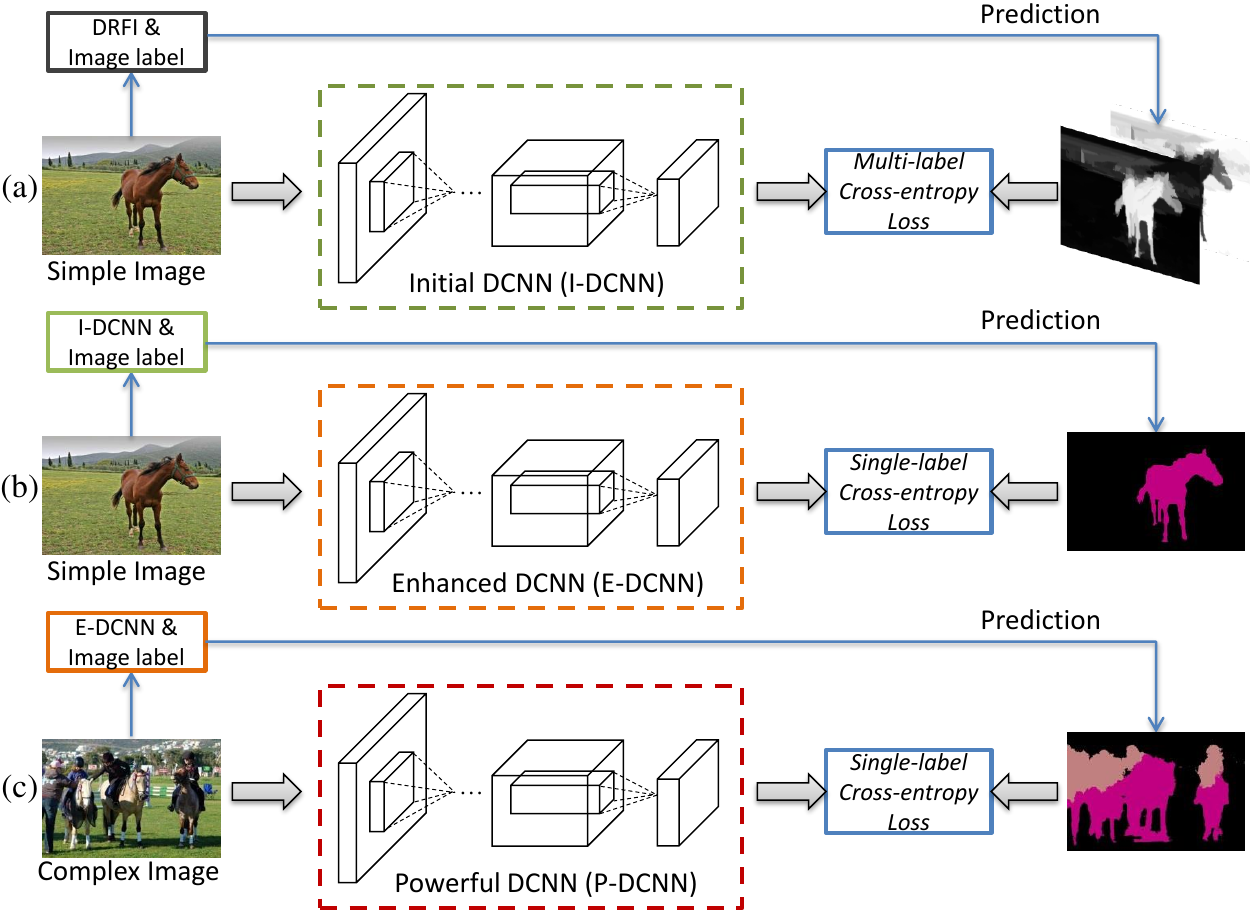}
	\caption{An illustration of the proposed simple to complex (STC) framework.
		(a) High quality saliency maps of simple images are first generated by DRFI~\cite{jiang2013salient} as
		the supervised foreground/background masks to train the Initial-DCNN using the proposed loss function.
		(b) Then, a better Enhanced-DCNN is learned, supervised with the segmentation masks predicted by Initial-DCNN. %
		(c) Finally, more masks of complex images are predicted to train a more powerful network, called Powerful-DCNN.}
	\label{fig:framework}
	\vspace{-1em}
\end{figure}
\vspace{-1em}
\section{Related Work}
\label{sec:related-work}

\subsection{Weakly Supervised Semantic Segmentation}
To reduce the burden of the pixel-level mask annotation, some weakly-supervised methods have been proposed for semantic segmentation.
Dai \emph{et al.}~\cite{2015-dai} and Papandreou \emph{et al.}~\cite{2015-papandreou-weakly} proposed to
estimate semantic segmentation masks by utilizing annotated bounding boxes.
For example, by incorporating pixel-level masks from the Pascal VOC~\cite{2010-pascal} and annotated
bounding boxes from the COCO~\cite{lin2014microsoft}, state-of-the-art results on PASCAL VOC
2012 benchmark were achieved by~\cite{2015-dai}.
To further reduce the burden of the bounding boxes collection,
some works~\cite{pathak2014fully,2015-papandreou-weakly,pinheiro2015weakly,pathak2015constrained,shimoda2016distinct,saleh2016built,kolesnikov2016seed} proposed to train the segmentation network by only using image-level labels.
Pathak \emph{et al.}~\cite{pathak2014fully} and Pinheiro \emph{et al.}~\cite{pinheiro2015weakly} proposed to
utilize multiple instance learning (MIL)~\cite{maron1998framework} framework to train the DCNN for segmentation.
In \cite{2015-papandreou-weakly}, an alternative training procedure based on Expectation-Maximization (EM)
algorithm was presented to dynamically predict foreground (with semantics)/background pixels.
Pathak \emph{et al.}~\cite{pathak2015constrained} introduced constrained convolutional neural
networks for weakly-supervised segmentation.
Specifically,  by utilizing object size as additional supervision, significant improvements were
made by~\cite{pathak2015constrained}.
Most recently, three kinds of loss functions, \ie, seeding, expansion and constrain-to-boundary, were leveraged in~\cite{kolesnikov2016seed} to train the segmentation network. Saleh \emph{et al.}~\cite{saleh2016built} also proposed a relevant approach using foreground/background prior for learning to segment, which is able to evidence the effectiveness of our framework.

\subsection{Self-paced Learning}
Our framework first learns from simple images and then applies the learned network to complex
ones, which is related to self-paced learning~\cite{kumar2010self}.
Recently, various computer vision applications~\cite{tang2012shifting,jiang2014self,liang2014computational}
based on self-paced learning have been proposed.
In specific, Tang \emph{et al.}~\cite{tang2012shifting} adapted object detectors learned from images to
videos by starting with easy samples.
Jiang \emph{et al.}~\cite{jiang2014self} addressed the data diversity.
In~\cite{liang2014computational}, very few samples were used as seeds to train a weak object detector,
and then more instances were iteratively accumulated to enhance the object detector, which can be considered
as a slightly-supervised self-paced learning method.
However, different from self-paced learning where each iteration automatically selects samples for training, 
the simple or complex samples are defined according to their appearance (\eg, single/multiple object(s) or 
clean/cluttered background) before training in this work.
\begin{figure*}[t]
	\centering
	\includegraphics[scale=0.72]{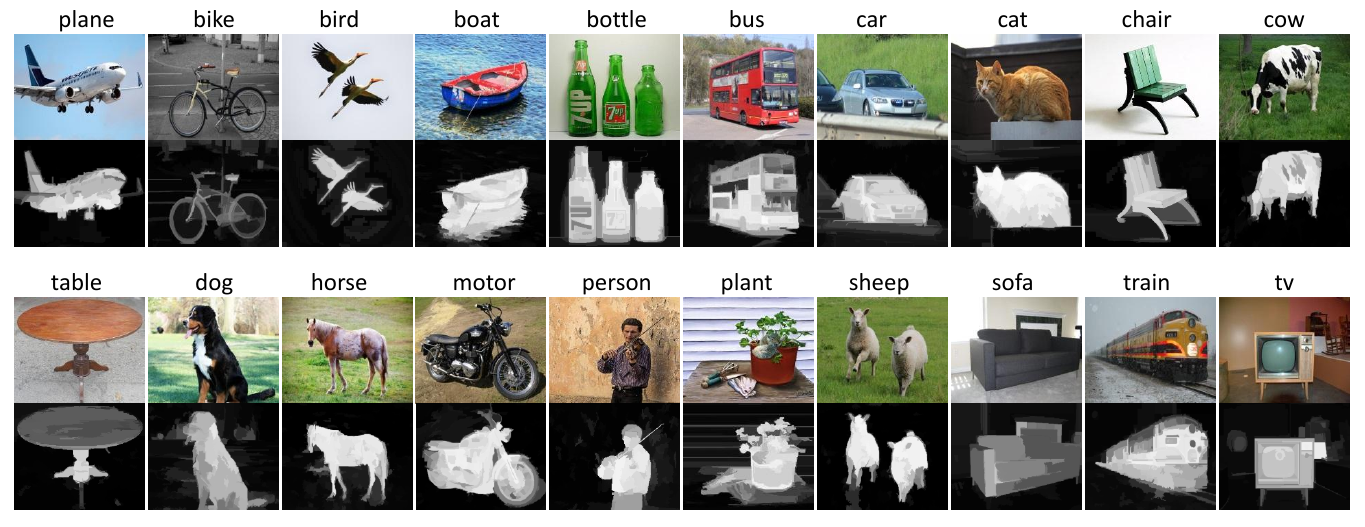}
	\caption{Examples of simple images and the corresponding saliency maps generated by DRFI on the 20 classes of PASCAL VOC.}
	\label{fig:sal_examp}
	\vspace{-1.5em}
\end{figure*}

Besides, many other works~\cite{xu2014tell,xu2015learning,liu2012weakly,rubinstein2012annotation,vezhnevets2012weakly} have also addressed this task. These methods are usually applied on simple or small scale datasets, \eg, MSRA~\cite{shotton2009textonboost} and SIFT-flow~\cite{liu2009nonparametric}. Specifically, Liu \emph{et al.}~\cite{liu2012weakly} proposed a graph propagation method to automatically assign the annotated labels at image level to those contextually derived semantic regions. Xu \emph{et al.}~\cite{xu2014tell} presented a latent structured prediction framework, where the graphical model encodes the presence and absence of a class as well as assignments of semantic labels to super-pixels. Vezhnevets \emph{et al.}~\cite{vezhnevets2012weakly} proposed a maximum expected agreement model selection principle that evaluates the quality of a model from the parametric family of structured models for semantic segmentation.
\vspace{-4mm}
\section{Proposed Method}
\label{sec:proposed-method}
Figure~\ref{fig:framework} shows the architecture of the proposed simple to complex (STC) framework.
We utilize the state-of-the-art saliency detection method, \ie, discriminative regional feature
integration (DRFI)~\cite{jiang2013salient}, to generate the saliency maps of simple images.
The produced saliency maps are first employed to train an initial DCNN with a multi-label cross-entropy loss function.
Then the simple to complex framework is proposed to gradually improve the capability of segmentation DCNN.

\vspace{-3mm}
\subsection{Initial-DCNN}
\label{ssec:smt}
For the generated saliency map of each image, the larger pixel value means it is more likely that this pixel belongs to foreground.
Figure \ref{fig:sal_examp} shows some instances of simple images and the corresponding 
saliency maps generated by DRFI. It can be observed that there exists explicit association 
between the foreground pixels and the semantic object(s).
Since each simple image is accompanied with a semantic label, it can be easily inferred that foreground
candidate pixels can be assigned with the corresponding image-level label.
Then, a multi-label cross-entropy loss function is proposed to train the segmentation network supervised by
saliency maps.

Suppose there are $C$ classes in the training set. We denote $\mathcal{O}_I = \{1, 2, \cdots, C\}$ and $\mathcal{O}_P = \{0, 1, 2, \cdots, C\}$ as the category sets for image-level label and pixel-level label, 
respectively, where $0$ indicates the $background$ class. 
We denote the segmentation network filtering by $f(\cdot)$, where all the convolutional layers filter the given
image $I$.
The $f(\cdot)$ produces a $h \times w \times (C+1)$ dimensional output of activations, where $h$ and $w$ are
the height and the width of the feature map for each channel, respectively.
We utilize the softmax function to compute the posterior probability of each pixel of $I$ belonging to
the $k^{th}$ $(k \in \mathcal{O}_P)$ class, which is formulated as follows,

\begin{equation}
\label{eq:soft}
{p_{ij}^k} = \frac{{\exp \left( {f_{ij}^k(I)} \right)}}{{\sum\limits_{l \in \mathcal{O}_P} {\exp \left( {f_{ij}^l(I)} \right)} }},
\end{equation}
where ${f_{ij}^k(I)}$ is the activation value at location $(i, j)$ $(1 \le i \le h,1 \le j \le w)$ of
the $k^{th}$ feature map.
In general, we define the probability obtained from the saliency map of the $l^{th}$ class at the location $(i, j)$ as
${{\hat p}^l}_{ij}$ ($\sum\limits_{l \in \mathcal{O}_P} {{{\hat p}^l}_{ij}}  = 1$).
Then, the multi-label cross-entropy loss function for semantic segmentation is then defined as

\begin{equation}
\label{eq:gloss}
 - \frac{1}{{h \times w}}\sum\limits_{i = 1}^h {\sum\limits_{j = 1}^w {\sum\limits_{l \in \mathcal{O}_P} {{{\hat p}^l}_{ij}\log ({p^l_{ij}})} } }.
\end{equation}

Specifically, for each simple image, we assume that only one semantic label is included.
Suppose that the simple image $I$ is annotated by the $c^{th}$ ($c \in \mathcal{O}_I$) class, and then the
normalized value from the saliency map is taken as the probability of each pixel belonging to the class $c$.
We resize the saliency map to the same size of the output feature map from the DCNN and Eqn.~(\ref{eq:gloss})
can then be re-formulated as
\begin{equation}
\begin{aligned}
 - \frac{1}{{h \times w}}\sum\limits_{i = 1}^h {\sum\limits_{j = 1}^w {({{\hat p}^c}_{ij}\log ({p^c_{ij}}) + {{\hat p}^{0}}_{ij}\log ({p^{0}_{ij}}))} },
\end{aligned}
\end{equation}
where ${p^{0}_{ij}}$ indicates the probability of the pixel at location $(i, j)$ belonging to the
$background$ (${p^{0}_{ij}} = 1 - {p^c_{ij}}$).
We denote the segmentation network learned in this stage as Initial-DCNN (I-DCNN for short).

It should be noted that we can also utilize SaliencyCut~\cite{cheng2015global} to generate the foreground/background
segmentation masks based on the generated saliency maps.
Then, single-label cross-entropy loss can be employed for training.
We compare this scheme with our proposed method, and find that the performance on VOC 2012 \emph{val} set will drop by 3\%.
The reason is that some saliency detection results are inaccurate.
Therefore, directly applying SaliencyCut~\cite{cheng2015global} to generate segmentation masks will introduce many nosies, which is
harmful for training the I-DCNN.
However, based on the proposed multi-label cross-entropy loss, correct semantic labels will still contribute to
the optimization, which can decrease the negative effect caused by low quality saliency maps.
\vspace{-3mm}
\subsection{Simple to Complex Framework}
In this section, a progressively training strategy is proposed by incorporating more complex images with
image-level labels to enhance the segmentation capability of DCNN.
Based on the trained I-DCNN, segmentation masks of images can be predicted, which can be used to further
improve the segmentation capability of DCNN.
Similar to the definition in Section~\ref{ssec:smt}, we denote the predicted probability for
the $k^{th}$ class at the location $(i, j)$ as $p^k_{ij}$.
Then, the estimated label $g_{ij}$ of the pixel at location $(i, j)$ by the segmentation DCNN can be formulated as
\begin{equation}
\label{eq:argmax}
g_{ij} = \arg \mathop {\max }\limits_{k \in \mathcal{O}_P} {p^k_{ij}}.
\end{equation}
\subsubsection{Enhanced-DCNN}
However, incorrect predictions from the I-DCNN may lead to the drift in semantic segmentation when used as
the supervision for training DCNN.
Fortunately, for each simple image in the training set, the image-level label is given, which can be utilized
to refine the predicted segmentation mask.
Specifically, if the simple image $I$ is labeled with $c$ ($c \in \mathcal{O}_I$), the estimated label of
the pixel can be re-formulated as
\begin{equation}
\label{eq:maxs}
g_{ij} = \arg  \max \limits_{k \in \{0, c\}} {p^{k}_{ij}},
\end{equation}
where $0$ indicates the category of $background$.
In this way, some false predictions for simple images in the training set can be eliminated.
Then, a more powerful segmentation DCNN called Enhanced-DCNN (E-DCNN for short) is trained by utilizing the
predicted segmentation masks as the supervised information.
We train the E-DCNN with the single-label cross-entropy loss function, which is widely used by
fully-supervised schemes~\cite{2015-long}.

\subsubsection{Powerful-DCNN}
In this stage, complex images with image-level labels, in which more semantic objects and cluttered background
are included, are utilized to train the segmentation DCNN.
Compared with I-DCNN, E-DCNN possesses a more powerful semantic segmentation capability due to the usage of the
large number of predicted segmentation masks.
Although E-DCNN is trained with simple images, the semantic objects in those images have large variety in
terms of appearance, scale and viewpoint, which is consistent with their appearance variation in complex images.
Therefore, we can apply E-DCNN to predict the segmentation masks of complex images.
Similar as Eqn. (\ref{eq:maxs}), to eliminate false predictions, the estimated label for each pixel of image $I$
is formulated as
\begin{equation}
g_{ij} = \arg \mathop {\max }\limits_{k \in \Omega } {p^k_{ij}},
\end{equation}
where $\Omega $ indicates the set of ground-truth semantic labels (including \emph{background}) for each image $I$.
We denote the segmentation network trained in this stage as Powerful-DCNN (P-DCNN for short).

\begin{figure}[t]
	\centering
	\includegraphics[scale=0.55]{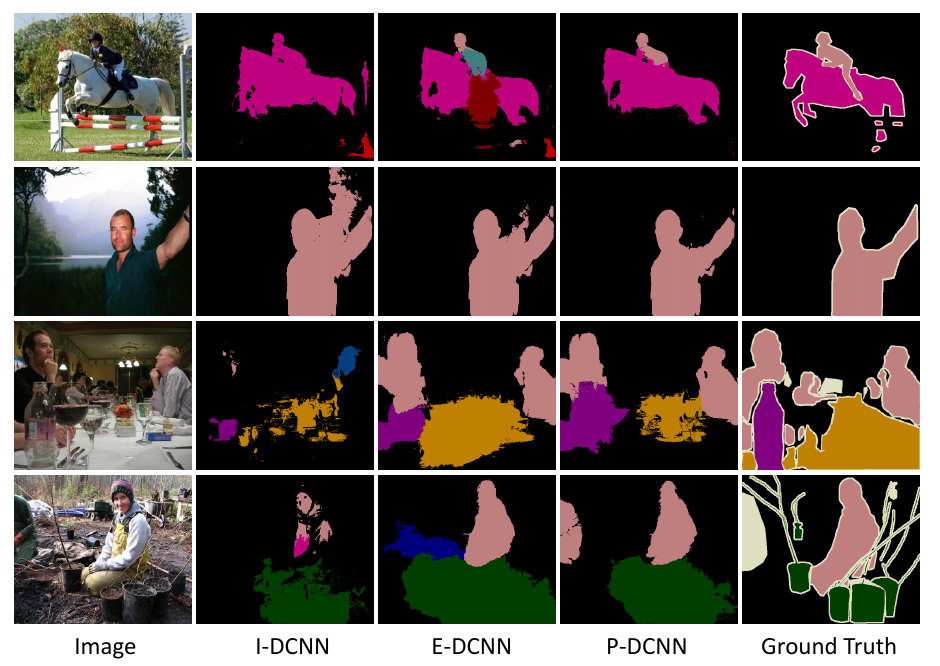}
	\caption{Examples of segmentation results generated by I-DCNN, E-DCNN and P-DCNN on the PASCAL VOC 2012 \emph{val} set, respectively.}
	\label{fig:stc_examp}
	\vspace{-1.8em}
\end{figure}

In this work, two kinds of cross-entropy losses are utilized to train segmentation networks. In particular, cross-entropy loss in the fully convolutional network is a pixel-wise one. For the fully supervised scheme, each pixel can only be assigned to one class and the corresponding cross-entropy is a single-label one. This matches the target of E-DCNN and P-DCNN. Therefore, we train these two networks using the single-label loss. For training the I-DCNN, the class information of each pixel can not be exactly obtained. To address this issue, each pixel is softly associated with two classes (one is background and the other is one of the 20 foreground classes) with different probabilities according to the produced saliency map and image-level label. We consider the loss function for this scheme as the multi-label cross-entropy loss.
To illustrate the effectiveness of each step, some segmentation results generated by I-DCNN, 
E-DCNN and P-DCNN are shown in Figure \ref{fig:stc_examp}. It can be seen that the segmentation 
results are progressively becoming better based on the proposed simple to complex framework.

\vspace{-3mm}

\section{Experimental Results}
\label{sec:experiments}
\subsection{Dataset}
\textbf{Flickr-Clean:} We construct a new dataset called \texttt{Flickr-Clean} to train the segmentation network
of I-DCNN.
The keywords, whose semantics are consistent with those from PASCAL VOC, are employed as queries to retrieve
images on the image hosting website \texttt{Flickr.com}.
We crawl images in the first several pages of searching results and use the state-of-the-art saliency
detection method, \ie, discriminative regional feature integration (DRFI), to generate the saliency maps of
the crawled images.
In order to ensure that the images are simple ones, we adopt a method similar to that proposed in \cite{chen2009sketch2photo,cheng2014salientshape} to filter the crawled images.
We measure the impreciseness and incompleteness of the SaliencyCut~\cite{cheng2015global} segmentation as in\cite{chen2009sketch2photo}. Denote the number of pixels of the given image $I$ as $N_I$ and the number of foreground pixels of the corresponding segmentation mask as $N_f$. We reserve those images whose foreground regions fit $0.3*N_{I}<N_{f}<0.5*N_{I}$. Without such a filtering scheme to clean up the training set, the performance of using all 100K crawled images for training results in 4\% performance drop for the I-DCNN. In the end, 41,625 images are collected to train the segmentation network.

\noindent \textbf{PASCAL VOC 2012:} The proposed weakly-supervised method is evaluated on the PASCAL
VOC 2012 segmentation benchmark~\cite{2010-pascal}. The original training data contain 1,464 images.
In~\cite{hariharan2011semantic}, 10,582 extra images ($train\_aug$) are annotated for training.
In our experiment, 10,582 images with only image-level labels are utilized as the complex image set for training.
The \emph{val} and \emph{test} sets have 1,449 and 1,456 images, respectively.
For both \emph{val} and \emph{test} sets, we only use the simple images from \texttt{Flickr-Clean} and the
complex images from $train\_aug$ for training.
The performance is measured in terms of pixel intersection-over-union (IoU) averaged on 21 classes
(20 \emph{object} and one \emph{background}).
Extensive evaluation of the weakly-supervised method is primarily conducted on the \emph{val} set and we also
report the result on the \emph{test} set (whose ground-truth masks are not released) by submitting the results to
the official PASCAL VOC 2012 server.

\vspace{-3mm}
\subsection{Training Strategies}
We employ the proposed simple to complex framework to learn the DCNN component of the
DeepLab-CRF model~\cite{chen2014semantic}, whose parameters are initialized by the
VGG-16 model~\cite{simonyan2014very} pre-trained on ImageNet~\cite{2009-imagenet}.
For the training of segmentation DCNNs (I-DCNN, E-DCNN and P-DCNN), we use a mini-batch size of 8 images.
Every training image is resized to $330 \times n$ and patches with the size of $321 \times 321$ are randomly
cropped during the training stage.
The initial learning rate is set as 0.001 (0.01 for the last layer) and divided by 10 after almost every 5 epochs.
The momentum and the weight decay are set as 0.9 and 0.0005.
The training progress is performed for about 15 epochs. To fairly compare our results with
those from~\cite{2015-papandreou-weakly,pathak2015constrained}, dense CRF inference is adopted to post-process the predicted results.
Each segmentation DCNN is trained based on a NVIDIA GeForce Titan GPU with 6GB memory.
All the experiments are conducted using DeepLab code~\cite{chen2014semantic}, which is implemented based on
the publicly available Caffe framework~\cite{jia2014caffe}.
\vspace{-3mm}
\subsection{Justifications}

\begin{table}
	\centering
	\caption{Comparison of I-DCNN models trained with different saliency maps on VOC 2012 \emph{val} set (mIoU in \%).}
	\begin{tabular}{c|c|c}
		\toprule
		Saliency Method & HS~\cite{shi2016hierarchical} & DRFI~\cite{jiang2013salient}  \\
		\midrule
		I-DCNN & 42.5  & 44.9 \\
		\bottomrule
	\end{tabular}	
	\label{tab:sal}	
	\vspace{-0.5em}
\end{table}
{\noindent \textbf{Justifications of Different Saliency Detecton Methods:} DRFI achieves the state-of-the-art performance on 6 popular benchmark datasets indicated by~\cite{2015-aliborji}. To investigate the quality of saliency maps generated by DRFI and how the performance of our proposed method varies with adopting different saliency detection methods, we train another I-DCNN model based on saliency maps produced by one of the latest methods, \ie, Hierarchical Saliency (HS)~\cite{shi2016hierarchical} detection method. Table~\ref{tab:sal} shows the segmentation results of I-DCNN models trained by different saliency maps. It can be seen that using DRFI saliency maps to train I-DCNN is effective.}
\begin{table}
	\centering
	\caption{Comparison of I-DCNN models trained on different numbers of images on VOC 2012 \emph{val} set (mIoU in \%).}
	\label{tab:size}
	\begin{tabular}{c|c|c|c|c|c}
		\toprule
		\texttt{Flickr-Clean} & 1/16 & 1/8 & 1/4 & 1/2 & All  \\
		\midrule
		I-DCNN & 39.8  & 42.1 & 45.7 & 45.6 & 44.9 \\
		\bottomrule
	\end{tabular}
	\vspace{-1em}
\end{table}

\noindent \textbf{Justifications of the Number of Training Images:} To investigate when the increasing number of collected training images will saturate the performance of the proposed method, we train I-DCNN models using varying numbers of images from \texttt{Flickr-Clean} dataset (see also Table 2). Each smaller set is a subset of the following larger set. For example, the 1/16 set is a subset of the 1/8 set. We firstly observe performance improvements when incorporating more training samples, which is quite intuitive. After getting best performance when using 1/4 of the training samples, further increasing the training samples hurts the performance. We believe the reason is that Flickr images ranked in the last few pages are quite noisy, and cannot be efficiently utilized by our current scheme. In this paper, our experiments are based on all images from \texttt{Flickr-Clean}.

\noindent\textbf{Justifications of the Simple to Complex Framework:} Table~\ref{tab:val-comp-net} shows the comparisons of different segmentation
DCNNs. It can be observed that, based on the proposed multi-label cross-entropy loss, saliency maps of simple images accompanied with image-level labels can be conveniently employed to train an effective neural network for semantic segmentation. The performance of I-DCNN is 44.9\%, which can outperform most state-of-the-arts. Besides, training with the segmentation masks predicted by I-DCNN can further improve the capability of semantic segmentation, \ie, 46.3\% \emph{vs.} 44.9\%. In addition, based on the enhanced neural network (\ie, E-DCNN), the performance can be further boosted, \ie, 49.8\% \emph{vs.} 46.3\%, by adding more complex images for training. Therefore, for the weakly-supervised semantic segmentation task, the proposed simple to complex (STC) framework is effective. In addition, we also conduct experiments of training I-DCNN with complex images to validate the necessity of using simple images. The mIoU score is 17.6\%, which is far below the result of ours. Please refer to the supplementary marital for more details.

\vspace{-3mm}
\subsection{Comparison with State-of-the-art Methods}
\begin{table}[]
	\centering
	\caption{Comparison of different segmentation DCNNs on VOC 2012 \emph{val} set.}
	\label{tab:val-comp-net}
	\begin{tabular}{c|c|c}
		\toprule
		Method & Training  Set & mIoU  \\
		\midrule
		I-DCNN & Flickr-Clean  & 44.9 \\
		E-DCNN & Flickr-Clean  & 46.3 \\
		P-DCNN & Flickr-Clean + VOC & 49.8 \\
		\bottomrule
	\end{tabular}
	\vspace{-1em}
\end{table}

\begin{table*}\setlength{\tabcolsep}{1.5pt}
	\centering 
	\caption{Comparison of weakly-supervised semantic segmentation methods on VOC 2012 \emph{val} set.}
	\label{tab:val-detail}
	\begin{tabular}{lcccccccccccccccccccccc}
		\toprule
		Methods &{bkg}&{plane}&{bike}&{bird}&{boat}&{bottle}&{bus}&{car}&{cat}&{chair}&{cow}&{table}&{dog}&{horse}&{motor}&{person}&{plant}&{sheep}&{sofa}&{train}&{tv}&  mIoU  \\
		\midrule
		\multicolumn{23}{l}{Other state-of-the-art methods:}  \\
		MIL-FCN~\cite{pathak2014fully} & - & -  & -  & -  & -  & -  & -  & -  & -  & -  & -  & -  & -  & -  & -  & -  & -  & -  & -  & -  & - & 25.7 \\
		EM-Adapt~\cite{2015-papandreou-weakly} & - & -  & -  & -  & -  & -  & -  & -  & -  & -  & -  & -  & -  & -  & -  & -  & -  & -  & -  & -  & - & 38.2 \\
		CCNN~\cite{pathak2015constrained}  &  65.9  &  23.8  &  17.6  &  22.8  &  19.4  &  36.2  &  47.3  &  46.9  &  47.0  &  $\bm{16.3}$  &  36.1  &  22.2  &  43.2  &  33.7  &  44.9  &  39.8  &  29.9  &  33.4  &  22.2  &  38.8  &  36.3  &  34.5  \\
		MIL*~\cite{pinheiro2015weakly} &  37.0  &  10.4  &  12.4  &  10.8  &  5.3  &  5.7  &  25.2  &  21.1  &  25.2  &  4.8  &  21.5  &  8.6  &  29.1  &  25.1  &  23.6  &  25.5  &  12.0  &  28.4  &  8.9  &  22.0  &  11.6  &  17.8  \\
		MIL-ILP*~\cite{pinheiro2015weakly}  &  73.2  &  25.4  &  18.2  &  22.7  &  21.5  &  28.6  &  39.5  &  44.7  &  46.6  &  11.9  &  40.4  &  11.8  &  45.6  &  40.1  &  35.5  &  35.2  &  20.8  &  41.7  &  17.0  &  34.7  &  30.4  &  32.6  \\
		MIL-ILP-sppxl*~\cite{pinheiro2015weakly}  &  77.2  &  37.3  &  18.4  &  25.4  &  28.2  &  31.9  &  41.6  &  48.1  &  50.7  &  12.7  &  45.7  &  14.6  &  50.9  &  44.1  &  39.2  &  37.9  &  28.3  &  44.0  &  19.6  &  37.6  &  35.0  &  36.6  \\
		MIL-ILP-bb*~\cite{pinheiro2015weakly}  &  78.6  &  46.9  &  18.6  &  27.9  &  30.7  &  38.4  &  44.0  &  49.6  &  49.8  &  11.6  &  44.7  &  14.6  &  50.4  &  44.7  &  40.8  &  38.5  &  26.0  &  45.0  &  20.5  &  36.9  &  34.8  &  37.8  \\
		MIL-ILP-seg*~\cite{pinheiro2015weakly}  &  79.6  &  50.2  &  $21.6$  &  40.6  &  34.9  &  40.5  &  45.9  &  51.5  &  60.6  &  12.6  &  51.2  &  11.6  &  56.8  &  52.9  &  44.8  &  42.7  &  31.2  &  55.4  &  21.5  &  38.8  &  $\bm{36.9}$  &  42.0  \\
		DCSM~\cite{shimoda2016distinct}  &  76.7  &  45.1  &  24.6  &  40.8  &  23.0  &  34.8  &  61.0  &  51.9  &  52.4  &  15.5  &  45.9  &  32.7  &  54.9  &  48.6  &  57.4  &  51.8  &  38.2  &  55.4  &  32.2  &  42.6  &  39.6  &  44.1  \\
		BFBP~\cite{saleh2016built}  &  79.2  &  60.1  &  20.4  &  50.7  &  41.2  &  46.3  &  62.6  &  49.2  &  62.3  &  13.3  &  49.7  &  38.1  &  58.4  &  49.0  &  57.0  &  48.2  &  27.8  &  55.1  &  29.6  &  54.6  &  26.6  &  46.6  \\
		SEC~\cite{kolesnikov2016seed}  &  82.4  &  62.9  &  26.4  &  61.6  &  27.6  &  38.1  &  66.6  &  62.7  &  75.2  &  22.1  &  53.5  &  28.3  &  65.8  &  57.8  &  62 3 &  52.5  &  32.5  &  62.6  &  32.1  &  45.4  &  45.3  &  50.7  \\
		\midrule
		\multicolumn{23}{l}{Ours:}  \\
		STC*  &  84.5  &  68.0  &  19.5  &  60.5  &  42.5  &  44.8  &  68.4 &  64.0  & 64.8  &  14.5  &  52.0  &  22.8  &  58.0  &  55.3  &  57.8  &  60.5  &  40.6  &  56.7  &  23.0  & 57.1  &  31.2  &  49.8  \\
		\bottomrule
	\end{tabular}
	\vspace{-0.5em}
\end{table*}

\begin{table*}\setlength{\tabcolsep}{1.5pt}
	\centering
	\caption{Comparison of fully- and weakly- supervised semantic segmentation methods on VOC 2012 \emph{test} set.}
	\label{tab:test-detail}
	\begin{tabular}{lcccccccccccccccccccccc}
		\toprule
		Methods &{bkg}&{plane}&{bike}&{bird}&{boat}&{bottle}&{bus}&{car}&{cat}&{chair}&{cow}&{table}&{dog}&{horse}&{motor}&{person}&{plant}&{sheep}&{sofa}&{train}&{tv}&  mIoU  \\
		\midrule
		\multicolumn{23}{l}{Fully Supervised:}  \\
		SDS~\cite{hariharan2014simultaneous}  &  86.3  &  63.3  &  25.7  &  63.0  &  39.8  &  59.2  &  70.9  &  61.4  &  54.9  &  16.8  &  45.0  &  48.2  &  50.5  &  51.0  &  57.7  &  63.3  &  31.8  &  58.7  &  31.2  &  55.7  &  48.5  &  51.6  \\
		FCN-8s~\cite{2015-long}  &  -  &  76.8  &  34.2  &  68.9  &  49.4  &  60.3  &  75.3  &  74.7  &  77.6  &  21.4  &  62.5  &  46.8  &  71.8  &  63.9  &  76.5  &  73.9  &  45.2  &  72.4  &  37.4  &  70.9  &  55.1  &  62.2  \\
		DeepLab-CRF~\cite{chen2014semantic}  &  92.1  &  78.4  &  33.1  &  78.2  &  55.6  &  65.3  & 81.3  &  75.5 &  78.6  & 25.3  &  69.2  &  52.7  &  75.2  &  69.0  &  79.1  &  77.6  &  54.7  &  78.3  &  45.1  &  73.3  &  56.2  &  66.4  \\
		\midrule
		\multicolumn{23}{l}{Weakly Supervised (other state-of-the-art methods):}  \\
		MIL-FCN~\cite{pathak2014fully} & - & -  & -  & -  & -  & -  & -  & -  & -  & -  & -  & -  & -  & -  & -  & -  & -  & -  & -  & -  & - & 24.9 \\
		EM-Adapt~\cite{2015-papandreou-weakly}  &  76.3  &  37.1  &  21.9  &  41.6  &  26.1  &  38.5  &  50.8  &  44.9  &  48.9  &  16.7  &  40.8  &  29.4  &  47.1  &  45.8  &  54.8  &  28.2  &  30.0  &  44.0  &  29.2  &  34.3  &  46.0  &  39.6  \\
		CCNN~\cite{pathak2015constrained}  &  -  &  21.3  &  17.7  &  22.8  &  17.9  &  38.3  &  51.3  &  43.9  &  51.4  &  15.6  &  38.4  &  17.4  &  46.5  &  38.6  &  53.3  &  40.6  &  34.3  &  36.8  &  20.1  &  32.9  &  38.0  &  35.5  \\
		MIL-ILP-sppxl*~\cite{pinheiro2015weakly}  &  74.7  &  38.8  &  19.8  &  27.5  &  21.7  &  32.8  &  40.0  &  50.1  &  47.1  &  7.2  &  44.8  &  15.8  &  49.4  &  47.3  &  36.6  &  36.4  &  24.3  &  44.5  &  21.0  &  31.5  &  41.3  &  35.8  \\
		MIL-ILP-bb*~\cite{pinheiro2015weakly}  &  76.2  &  42.8  &  20.9  &  29.6  &  25.9  &  38.5  &  40.6  &  51.7  &  49.0  &  9.1  &  43.5  &  16.2  &  50.1  &  46.0  &  35.8  &  38.0  &  22.1  &  44.5  &  22.4  &  30.8  &  43.0  &  37.0  \\
		MIL-ILP-seg*~\cite{pinheiro2015weakly}  &  78.7  &  48.0  &  21.2  &  31.1  &  28.4  &  35.1  &  51.4  &  55.5  &  52.8  &  7.8  &  56.2  &  19.9  &  53.8  &  50.3  &  40.0  &  38.6  &  27.8  &  51.8  &  24.7  &  33.3  &  46.3  &  40.6  \\
		DCSM~\cite{shimoda2016distinct}  &  78.1  &  43.8  &  26.3  &  49.8  &  19.5  &  40.3  &  61.6  &  53.9  &  52.7  &  13.7  &  47.3  &  34.8  &  50.3  &  48.9  &  69.0  &  49.7  &  38.4  &  57.1  &  34.0  &  38.0  &  40.0  &  45.1  \\
		BFBP~\cite{saleh2016built}  &  80.3  &  57.5  &  24.1  &  66.9  &  31.7  &  43.0  &  67.5  &  48.6  &  56.7  &  12.6  &  50.9  &  42.6  &  59.4  &  52.9  &  65.0  &  44.8  &   41.3  &   51.1  &   33.7  &   44.4  &   33.2  &   48.0  \\
		SEC~\cite{kolesnikov2016seed}  &  83.5  &  56.4  &  28.5  &  64.1  &  23.6  &  46.5  &  70.6  &  58.5  &  71.3  &  23.2  &  54.0  &  28.0  &  68.1  &  62.1  &  70.0 &  55.0  &  38.4  &  58.0  &  39.9  &  38.4  &  48.3  &  51.7  \\
		\midrule
		\multicolumn{23}{l}{Weakly Supervised (ours):}  \\
		STC*  &  85.2  &  62.7  &  21.1  &  58.0  &  31.4  &  55.0  &  68.8  &  63.9  &  63.7  &  14.2  &  57.6  &  28.3  &  63.0  &  59.8  &  67.6  &  61.7  &  42.9  &  61.0  &  23.2  &  52.4  &  33.1  &  51.2  \\
		\bottomrule
	\end{tabular}
	\vspace{-0.5em}
\end{table*}

Table~\ref{tab:val-detail} shows the detailed results of
ours compared with those of state-of-the-art methods.
* indicates those methods that use additional images to train the segmentation network.
For MIL-FCN~\cite{pathak2014fully}, EM-Adapt~\cite{2015-papandreou-weakly}, CCNN~\cite{pathak2015constrained}, DCSM~\cite{shimoda2016distinct}, BFBP~\cite{saleh2016built} and SEC~\cite{kolesnikov2016seed},
the segmentation networks are trained on $train\_aug$ taken from VOC 2012.
For MIL-ILP-*~\cite{pinheiro2015weakly}, the segmentation network is trained with 700K images for 21 classes taken
from ILSVRC 2013.
Image-level prior (ILP), and some smooth priors, \ie, superpixels (-sppxl), BING~\cite{cheng2014bing} boxes (-bb) and MCG~\cite{arbelaez2014multiscale}
segmentations (-seg), are utilized for post-processing to further boost the segmentation results.
The proposed framework is learned on 50K (40K simple images from \texttt{Flickr-Clean} and 10K complex images from PASCAL VOC) images, which are much fewer compared with those of~\cite{pinheiro2015weakly} (700K).
Surprisingly, our result can make a significant improvement compared with the best 
result of~\cite{pinheiro2015weakly} (49.8\% \emph{vs.} 42.0\%). It can be observed that SEC~\cite{kolesnikov2016seed} achieves the state-of-the-art performance on this challenging task. The superiority of SEC mainly benefits from using CRF-based constrain-to-boundary loss for network optimizing. By only using cross-entropy loss, the mIoU score reported in~\cite{kolesnikov2016seed} is 45.4\%. Based on simple images that are cheap to obtain, our STC framework can easily achieve the competitive performance (49.8\% \emph{vs.} 50.7\%) by simply employing cross-entropy loss.

Table~\ref{tab:test-detail} reports our results on PASCAL
VOC 2012 \emph{test} set and compare them with the state-of-the-art weakly-supervised methods.
It can be observed that our result is competitive compared with the state-of-the-art performance (51.2\% \emph{vs.} 51.7\%).
For EM-Adapt~\cite{2015-papandreou-weakly}, the segmentation network is learned based on $train\_aug$ and $val$ sets.
In~\cite{pathak2015constrained}, by adding additional supervision of object size information, the performance can
be improved from 35.5\% to 45.1\%.
We also compare our result with several fully-supervised methods in Table~\ref{tab:test-detail}.
It can be observed that we have made a significant improvement to approach those results learned with fully supervised schemes.
In particular, our weakly-supervised framework achieves similar results compared with SDS~\cite{hariharan2014simultaneous}, which is learned in a fully supervised manner.
Besides, we conduct additional experiments based on the semi-supervised setting. The experimental results demonstrate that STC can also boost the segmentation performance when only a small number of fully-supervised images is available. More detailed comparative analyses are provided in the supplementary material.

\begin{figure*}
	\centering
	\includegraphics[scale=0.832]{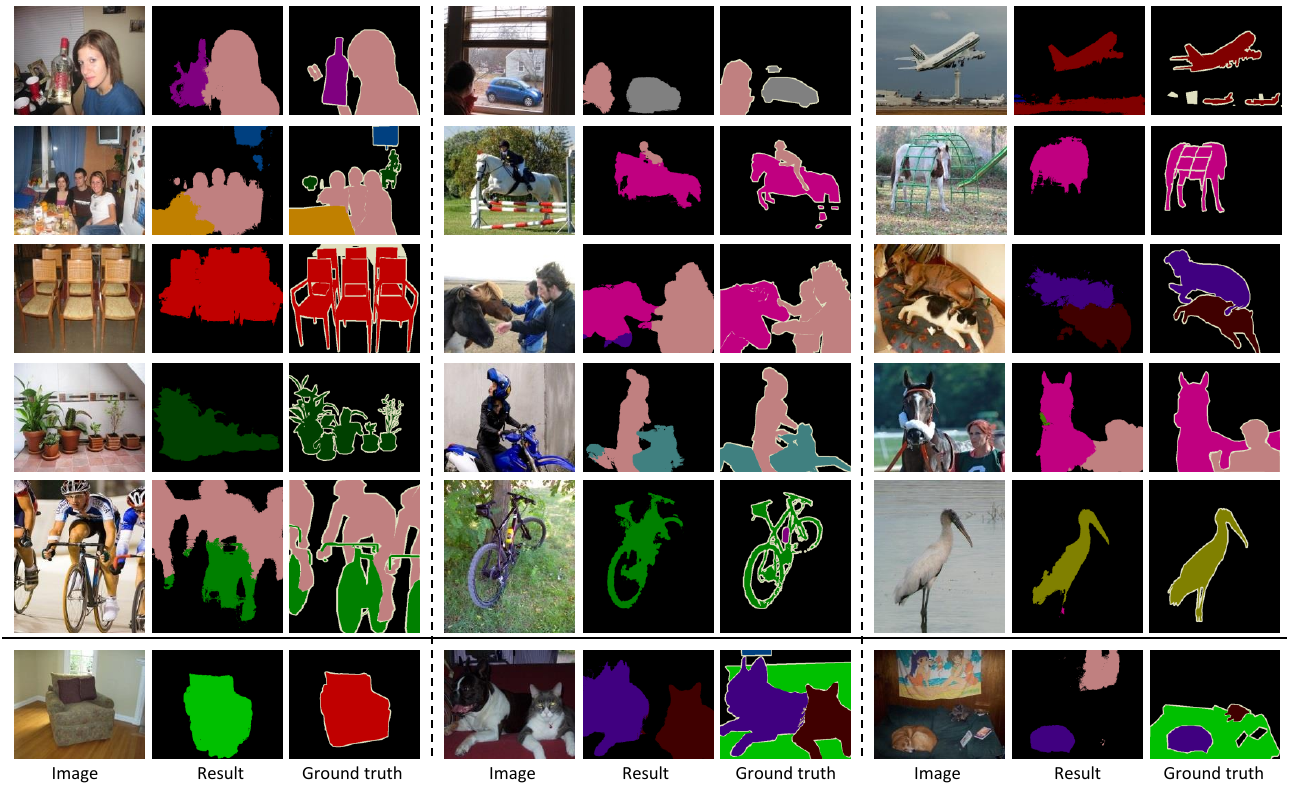}
	\caption{Qualitative segmentation results on PASCAL VOC 2012 \emph{val} set. Some failure cases are shown in the last row.}
	\label{fig:example}
	\vspace{-1em}
\end{figure*}

Qualitative segmentation results obtained by the proposed framework are shown in 
Figure~\ref{fig:example}. Some failure cases are shown in the last row of Figure~\ref{fig:example}.
In the first case 
(row: 6, column: 1), the \emph{chair} has a similar appearance as \emph{sofa} and the pixels of 
foreground segmentation are totally predicted as \emph{sofa}. In the second (row: 6, 
column: 2) and the third case (row: 6, column: 3), \emph{sofa} which occupies a large 
region of the image is wrongly predicted as \emph{background}. Including more samples with clean background and various appearances for training or using classification results for post-processing 
may help solve these issues.

\vspace{-3mm}
\subsection{Discussion}
The comparison between the proposed STC and \cite{pinheiro2015weakly} is a little unfair. 
The deep neural network utilized in \cite{pinheiro2015weakly} is based on \texttt{OverFeat} \cite{sermanet2013overfeat}, 
in which there are 10 weight layers, 
while in this paper, we utilize the VGG-16 model, which has 16 weight layers, as the basic architecture of the segmentation network. 
Both two models are pre-trained on ImageNet and the VGG-16 model works better than the \texttt{OverFeat} model 
on the ILSVRC~\cite{2009-imagenet} classification task.
However, Pinheiro \emph{et al.} \cite{pinheiro2015weakly} utilized 700K images with image-level labels for 
training, which is a much larger number compared with the training set (50K) of ours.
In addition, the performance of \cite{pinheiro2015weakly} highly depends on complex post-processing. Without 
any post-processing step, the performance of \cite{pinheiro2015weakly} is 17.8\%, which is far below the result of ours, \ie, 49.8\%.
\vspace{-3mm}
%

%

%

\ifCLASSOPTIONcompsoc
  \section*{Acknowledgments}
  This work was sponsored by the National Key Research and Development of China (NO. 2016YFB0800404), the National Natural Science Foundation of China (NO. 61532005, NO. 61210006, NO. 61402268, NO. 61572264), and CAST young talents plan.
\else
  \section*{Acknowledgment}
  
\fi

\ifCLASSOPTIONcaptionsoff
  \newpage
\fi

\bibliographystyle{IEEEtran}
\bibliography{mybibfile}

\end{document}